\documentclass{article}

\usepackage[final,dblblindworkshop]{neurips_2026}
\workshoptitle{SocialAgent: Second Workshop on Large Language Models for Social Reasoning and Simulation}
\usepackage[utf8]{inputenc}
\usepackage[T1]{fontenc}
\usepackage{hyperref}
\usepackage{url}
\usepackage{booktabs}
\usepackage{amsmath}
\usepackage{microtype}
\usepackage{xcolor}
\usepackage{graphicx}
\usepackage{adjustbox}
\usepackage{float}
\usepackage{tikz}
\usetikzlibrary{arrows.meta,positioning,calc,backgrounds,fit,shapes.geometric}

\definecolor{dtblue}{HTML}{2563EB}
\definecolor{dtbluefill}{HTML}{EAF2FF}
\definecolor{dtpurple}{HTML}{7C3AED}
\definecolor{dtpurplefill}{HTML}{F2EBFF}
\definecolor{dtorange}{HTML}{D97706}
\definecolor{dtorangefill}{HTML}{FFF3DF}
\definecolor{dtgreen}{HTML}{059669}
\definecolor{dtgreenfill}{HTML}{E8F8F1}
\definecolor{dtink}{HTML}{243247}
\definecolor{dtmuted}{HTML}{64748B}

\title{From Survey Histories to Interpretable Digital Twins:\\Auto-Discovering Task-Specific Personas and Behavioral Mechanisms}

\author{Tianze Deng$^\ast$ \quad Iris Ye$^\ast$ \quad Ozan Candogan$^\dagger$\\
{\footnotesize $^\ast$Equal contribution. \quad $^\dagger$Faculty advisor.}}

\newcommand{\raw}{\textsc{Raw}}
\newcommand{\cmech}{\textsc{Channel~+~Mechanism}}

\begin{document}
\maketitle
{\makeatletter\renewcommand{\@makefntext}[1]{\noindent#1}\makeatother\begin{NoHyper}\footnotetext{Booth School of Business, University of Chicago.\\Contact: \texttt{\{tianze10,iris.ye,Ozan.Candogan\}@chicagobooth.edu}.}\end{NoHyper}}

\begin{abstract}
LLM-based ``digital twins'' aim to simulate how an individual would behave in new environments or respond to novel questions, given some representation of that individual’s prior responses. A common approach constructs this representation from survey transcripts or summaries and evaluates fidelity by predicting holdout responses. Prior work shows that compressing long transcripts into shorter LLM-generated summaries does not significantly reduce predictive accuracy, suggesting that information volume is not the primary bottleneck. In this work, we ask whether long survey histories can instead be transformed into compact, task-specific personas that remain predictive for unseen respondents and products within the same task while making simulated decisions more interpretable. We introduce an iterative auto-discovery pipeline that constructs persona representations and learns how persona evidence influences simulated responses. We evaluate the procedure on the 40-item Product Preferences--Pricing task of Twin-2K-500~\citep{toubia2025twin}, learning from 20 respondents and 20 products and evaluating the frozen representation and mechanism on 30 unseen respondents and 20 unseen products. The holdout answers come from two survey waves, and every prediction is scored against each wave separately. For \texttt{gpt-5.4-nano}, replacing the full survey transcript with the learned channel representation and mechanism improves exact accuracy from 0.573 to 0.623 and from 0.558 to 0.626 in the two waves. Larger and cross-provider simulators (\texttt{gpt-5.4}, \texttt{gpt-5-mini}, \texttt{qwen3-8b}) using the full transcript do not achieve statistically significant improvements over the baseline in both waves, and none significantly outperforms the learned representation in either wave. This pattern shows that the learned representation produces consistent gains against human responses collected in both waves while retaining an explicit mapping from respondent evidence to predicted decisions.

\end{abstract}

\section{Introduction}

LLM-based digital twins aim to simulate how an individual would behave or respond in new settings, given a representation of their prior data. A common instantiation elicits a long survey from each respondent, passes the resulting transcript to a language model, and asks it to predict holdout responses as that respondent would. The present study examines this survey-grounded individual-prediction setting using Twin-2K-500~\citep{toubia2025twin}, which pairs 500 input questions with 88 holdout questions across 17 distinct prediction tasks for over two thousand respondents and offers substantially broader per-individual coverage than prior digital-twin benchmarks. Broader coverage does not by itself yield a faithful twin. For individual-level simulation, the representation built from this transcript must preserve behaviorally relevant differences across respondents rather than collapse them into a generic profile, and evidence of weak human--twin correlation and systematic distortions indicates that additional personal context alone is insufficient~\citep{peng2025funhouse}. This suggests that the bottleneck is not simply information volume but how that information is represented and used, specifically which prior answers matter for a target question and how they should affect its prediction.

In many existing approaches, the mapping from a respondent's prior answers to a target prediction remains implicit. Approaches that supply respondent information directly as model context, and those that adapt the simulator to observed response distributions, both leave unspecified how specific respondent evidence shapes an individual prediction (Section~\ref{sec:related}). We instead make this mapping explicit at the level of which survey evidence enters an individual prediction, in which contexts, and in which direction. We refer to the items whose responses are to be predicted as \emph{target questions} and to the remainder of a respondent's survey history as \emph{non-target questions}. The pipeline reads the target questions themselves, not their answers, to extract the constructs and decision demands a prediction may depend on, and it then searches each respondent's non-target answers for the corresponding personal evidence. Our iterative pipeline learns both how to organize this evidence into a compact persona representation and how to use it when making a prediction. It separates the problem into three steps: identifying task-relevant behavioral dimensions, finding respondent-specific evidence for those dimensions, and learning when and in which direction that evidence should affect the predicted response. This decomposition provides \emph{decision-level interpretability} by making explicit which respondent evidence is used, in what question context, and how it influences the prediction. The interpretation concerns the simulator's observable input--output behavior rather than its internal computation or a causal psychological process in the respondent.

We test the procedure on the Product Preferences--Pricing block of Twin-2K-500, which we refer to as the \emph{pricing task}. Each of its 40 items describes one product, displays a price, and asks for a binary purchase decision (Figure~\ref{fig:pricing-input-example}). From labels in a single learning cell---20 development respondents crossed with 20 calibration products---the pipeline discovers two channels, spending restraint and brand distinctiveness, and a separate-routing mechanism. These are frozen and evaluated on 30 unseen respondents crossed with 20 unseen products. Each of the 600 cases carries human answers from two survey waves and is scored against each separately (Section~\ref{sec:pricing} details the label-blind split). For \texttt{gpt-5.4-nano}, replacing the full survey transcript with the learned representation raises exact accuracy from 0.573 to 0.623 in one wave and from 0.558 to 0.626 in the other, with paired intervals excluding zero in both. No tested comparator reading the full transcript improves over the \texttt{gpt-5.4-nano} transcript baseline in both waves, and none detectably exceeds the learned representation in either wave. The cross-wave consistency distinguishes the learned representation from these model-capacity substitutions without establishing model equivalence or superiority.

The contribution has three parts. First, we introduce an auto-discovery pipeline that jointly learns compact, task-specific persona representations and executable decision rules specifying when, in which direction, and under what conditions persona evidence should influence individual predictions. Second, we evaluate the frozen system under a leakage-controlled double holdout of unseen respondents and products across two survey waves. Third, the learned representation exposes an auditable evidence–context–prediction mapping, providing decision-level interpretability of simulated responses. Accordingly, the evidence supports transfer to unseen respondents and products within the pricing task, rather than generalization across survey constructs or task formats.

\section{Related work}
\label{sec:related}

\subsection{LLM-based human simulation and adaptation}
Prior work conditions language models on demographic profiles to reproduce opinion distributions~\citep{argyle2023out}, tests whether they replicate behavioral experiments~\citep{aher2023simulate}, builds interview-grounded agents for individual prediction~\citep{park2024generative}, and quantifies uncertainty in simulated population statistics~\citep{huang2025uncertainty}. Twin-2K-500 extends this setting to individual digital twins, with more than 500 responses for 2,058 participants and selected tasks repeated in a later wave~\citep{toubia2025twin}.

One route to improving human simulation is model post-training. Aggregate-distribution supervision adapts model parameters to population or subpopulation response distributions. \citet{cao2025specializing} fine-tune against country-level response distributions and evaluate transfer to unseen questions, countries, and a separate survey, whereas \citet{suh2025finetuning} fine-tune on 3{,}362 questions and 70{,}000 subpopulation--response pairs and reduce the predicted--human distribution gap by up to 46\%. Individual-response supervision instead trains on person-level observations. \citet{kolluri2025finetuning} fine-tune on 2.9 million individual responses from 400{,}491 participants across 210 social-science experiments, although their principal evaluation aggregates predictions into response distributions across experimental conditions.

A second route keeps model weights fixed and uses evaluation feedback to optimize prompts. OPRO supplies candidate instructions and their scores to a language model that proposes improved instructions~\citep{yang2024opro}, whereas GEPA reflects on system trajectories to diagnose failures and evolve one or more prompts~\citep{agrawal2026gepa}. The present study also keeps simulator weights fixed, but calibration feedback searches over a different object. Rather than optimizing only a general task instruction, the pipeline searches over a structured, survey-grounded representation comprising behavioral channels, respondent-specific evidence allocations, and decision rules. These components are frozen before evaluation on unseen respondents and products. Thus, model post-training, general prompt optimization, and the structured representation search studied here modify different parts of the simulation pipeline.

\subsection{Automated behavioral and model discovery}

A related line of work uses language models to automate model and theory discovery. BoxLM proposes executable probabilistic models, evaluates their fit, and uses model criticism to guide subsequent proposals~\citep{li2024boxlm}. AutoCog proposes discriminating experiments, collects human behavioral data, arbitrates between executable cognitive theories, and revises the weaker theory across cycles~\citep{jagadish2026autocog}. Causal Alignment uses within-experiment A/B comparisons as directional supervision for improving marketing content~\citep{angelopoulos2024causal}. These systems illustrate a closed-loop pattern in which candidate rules or models are proposed, evaluated against observed outcomes, and revised using the resulting feedback.

Our setting differs in what must be discovered before such a rule can be applied. The survey instrument constrains the available respondent evidence, but the relevant inputs to the prediction rule are not specified in advance. The pipeline must determine which behavioral dimensions are relevant to the target questions, recover respondent-specific evidence for those dimensions from non-target survey answers, and learn how that evidence should influence individual predictions. 

\section{Auto-discovery pipeline}
\label{sec:pipeline}

\begin{figure}[!tbp]
  \centering
  \resizebox{\linewidth}{!}{%
  \begin{tikzpicture}[
    font=\sffamily,
    object/.style={rounded corners=3pt, minimum height=2.65cm,
      text width=2.20cm, inner xsep=5pt, inner ysep=5pt,
      align=center, line width=0.75pt,
      execute at begin node={\hyphenpenalty=10000\exhyphenpenalty=10000}},
    decision/.style={diamond, aspect=1.35, draw=dtmuted!80, fill=white,
      text width=1.42cm, inner sep=1.5pt, align=center, line width=0.75pt,
      font=\scriptsize\bfseries, text=dtink},
    flow/.style={line width=0.85pt, draw=dtmuted,
      shorten <=1.0pt, shorten >=1.0pt},
    loop/.style={-{Latex[length=2.0mm,width=1.6mm]}, line width=0.8pt,
      dashed, rounded corners=4pt}
  ]
    \node[object, draw=dtblue, fill=dtbluefill, text=dtink] (layer1) at (1.15,3.70)
      {\textbf{LAYER 1}\\[2pt]
       \textbf{Channel proposal}\\[3pt]
       {\scriptsize\textbf{Input:} target-question text; target answers hidden}\\[3pt]
       {\scriptsize\textbf{Operation:} propose constructs and candidate source questions}\\[3pt]
       {\scriptsize\textbf{Output:}\\candidate channels}};
    \node[object, draw=dtorange, fill=dtorangefill, text=dtink] (layer2) at (3.95,3.70)
      {\textbf{LAYER 2}\\[2pt]
       \textbf{Evidence mapping}\\[-1pt]
       \textbf{and validation}\\[3pt]
       {\scriptsize\textbf{Input:} non-target persona Q--A}\\[3pt]
       {\scriptsize\textbf{Operation:} synthesize evidence and test ON vs. OFF}\\[3pt]
       {\scriptsize\textbf{Output:} validated channel--evidence mappings}};
    \node[decision] (channelif) at (6.45,3.70)
      {Any channel validated?};
    \node[object, draw=dtpurple, fill=dtpurplefill, text=dtink] (layer3) at (8.85,3.70)
      {\textbf{LAYER 3}\\[2pt]
       \textbf{Channel combination}\\[-1pt]
       \textbf{and mechanism}\\[3pt]
       {\scriptsize\textbf{Input:} validated channels and evidence}\\[3pt]
       {\scriptsize\textbf{Operation:} compare combination, routing, interaction, and scope}\\[3pt]
       {\scriptsize\textbf{Output:} selected mechanism}};
    \node[decision] (mechanismif) at (11.30,3.70)
      {Stable calibration gain?};
    \node[object, draw=dtgreen, fill=dtgreenfill, text=dtink] (holdout) at (13.65,3.70)
      {\textbf{FREEZE}\\[2pt]
       \textbf{Final prompt}\\[3pt]
       {\scriptsize validated channels\\$+$ persona evidence\\$+$ mechanism}\\[5pt]
       \textbf{DOUBLE-}\\\textbf{HOLDOUT}\\\textbf{EVALUATION}\\[2pt]
       {\scriptsize unseen personas $\times$ unseen products}};

    \draw[flow] (layer1.east) -- (layer2.west);
    \fill[dtmuted]
      ($(layer2.west)+(-0.16,0.10)$) --
      ($(layer2.west)+(-0.16,-0.10)$) --
      (layer2.west) -- cycle;
    \draw[flow] (layer2.east) -- (channelif.west);
    \fill[dtmuted]
      ($(channelif.west)+(-0.16,0.10)$) --
      ($(channelif.west)+(-0.16,-0.10)$) --
      (channelif.west) -- cycle;
    \draw[line width=0.85pt, draw=dtpurple, shorten <=1.0pt, shorten >=1.0pt]
      (channelif.east) --
      node[above, font=\scriptsize\bfseries, text=dtpurple] {YES}
      (layer3.west);
    \fill[dtpurple]
      ($(layer3.west)+(-0.16,0.10)$) --
      ($(layer3.west)+(-0.16,-0.10)$) --
      (layer3.west) -- cycle;
    \draw[flow] (layer3.east) -- (mechanismif.west);
    \fill[dtmuted]
      ($(mechanismif.west)+(-0.16,0.10)$) --
      ($(mechanismif.west)+(-0.16,-0.10)$) --
      (mechanismif.west) -- cycle;
    \draw[line width=0.85pt, draw=dtgreen, shorten <=1.0pt, shorten >=1.0pt]
      (mechanismif.east) --
      node[above, font=\scriptsize\bfseries, text=dtgreen] {YES}
      (holdout.west);
    \fill[dtgreen]
      ($(holdout.west)+(-0.16,0.10)$) --
      ($(holdout.west)+(-0.16,-0.10)$) --
      (holdout.west) -- cycle;

    \coordinate (channelreturn) at (6.45,0.20);
    \draw[loop, draw=dtblue] (channelif.south) -- (channelreturn) -| (layer1.south);
    \node[font=\scriptsize\bfseries, text=dtblue, fill=white, inner sep=1.5pt]
      at (3.80,-0.02) {NO --- next round: re-propose channels};
    \coordinate (mechanismreturn) at (11.30,0.62);
    \draw[loop, draw=dtpurple] (mechanismif.south) -- (mechanismreturn) -| (layer3.south);
    \node[font=\scriptsize\bfseries, text=dtpurple, fill=white, inner sep=1.5pt]
      at (10.05,0.39) {NO --- next round: refine mechanism};
  \end{tikzpicture}}
  \caption{\textbf{Auto-discovery pipeline.} Layer~1 proposes channels,
  and Layer~2 instantiates and validates their persona-specific evidence mappings.
  If none is supported, the next round returns to Layer~1. Otherwise Layer~3 keeps
  the validated representation and compares channel-combination and mechanism
  candidates. An unstable candidate triggers another Layer~3 round, whereas a stable
  candidate is frozen before the double-holdout evaluation.}
  \label{fig:pipeline}
\end{figure}

The pipeline separates three learned layers from the rounds that revise them. Figure~\ref{fig:pipeline} shows how the rounds move between or revisit layers, and Figure~\ref{fig:calibration-trace} details the trajectory realized in the pricing study.

\paragraph{Layer 1: channel skeleton.}
An LLM reads the study
description and the texts of the target questions. It then proposes candidate behavioral channels: task-specific constructs,
such as spending restraint, that may shape the target decisions.

\paragraph{Layer 2: evidence allocation and validation.}
For each development persona and each candidate channel, an allocator locates the persona's answers to the source questions named in Layer~1 and synthesizes them into that persona's channel evidence, preserving counterevidence and uncertainty. Layer~2 then validates each channel with an item-local ablation: for one development persona and one calibration item at a time, it scores the simulator's prediction with the channel's evidence added (ON) versus withheld (OFF), everything else held fixed. Let $\mathcal{P}_{\mathrm{dev}}$ denote the development personas, $\mathcal{Q}_{\mathrm{cal}}$ the calibration items, $\mathcal{C}$ the candidate channels, and $\mathcal{Q}_c\subseteq\mathcal{Q}_{\mathrm{cal}}$ the calibration items admitted by the scope proposed for channel $c$. For persona $p\in\mathcal{P}_{\mathrm{dev}}$, item $q\in\mathcal{Q}_c$, and channel $c\in\mathcal{C}$, define
\begin{equation}
  \begin{aligned}
    \delta_{p,q,c}
    &=
    \bar{s}\!\left(\hat{y}^{\mathrm{on}(c)}_{p,q},y_{p,q}\right)
    -
    \bar{s}\!\left(\hat{y}^{\mathrm{off}(c)}_{p,q},y_{p,q}\right),\\
    \Delta_c
    &=\frac{1}{|\mathcal{P}_{\mathrm{dev}}||\mathcal{Q}_c|}
    \sum_{q\in\mathcal{Q}_c}
    \sum_{p\in\mathcal{P}_{\mathrm{dev}}}\delta_{p,q,c}.
  \end{aligned}
  \label{eq:item-local-effect}
\end{equation}
The prediction $\hat{y}^{\mathrm{on}(c)}_{p,q}$ includes channel $c$'s evidence and $\hat{y}^{\mathrm{off}(c)}_{p,q}$ withholds it, $y_{p,q}$ is the respondent's answer, $s$ scores a prediction against it, and the bar averages repeated simulator runs. $\Delta_c$ is thus the average accuracy effect of adding channel $c$'s evidence over the development--calibration cell. For the binary pricing task, $s(\hat{y},y)=\mathbf{1}\{\hat{y}=y\}$, $\mathcal{Q}_c$ contains all 20 calibration products, and the OFF score is averaged over two byte-identical runs of the OFF condition. Round~1 retains a channel only if its evidence improves average accuracy, $\Delta_c>0$. If no channel satisfies this retention rule, the next round returns to Layer~1 to propose a revised channel skeleton. Otherwise the retained channel--evidence mappings advance to Layer~3.

\paragraph{Layer 3: executable consumer mechanism.}
The final layer compiles the validated channels into an executable consumer mechanism: a textual rule specifying when each channel applies, in which direction it shifts the prediction, and where it should not be used (Appendix~\ref{app:frozen-pricing-artifact} renders the frozen instance). Figure~\ref{fig:case-trace} illustrates how this frozen mechanism determines which persona-specific channel evidence is relevant for one holdout product. The resulting mechanism and channel evidence are then supplied jointly to the simulator.

Layer~3 searches over mechanisms for combining the retained channels. One class is \emph{separate routing}, in which each channel has its own \emph{context gate} determined by the current product context—for example, whether the product or brand is meaningfully characterized as mainstream or distinctive—and can influence the prediction independently of the other channels when that gate is satisfied. Layer~3 may also propose \emph{interaction} mechanisms, in which the effect of one channel depends on the joint state of one or more other channels. Such candidates can be introduced when heterogeneous item-local effects observed in Layer~2 suggest a possible cross-channel dependence. Mechanism candidates are evaluated by aggregate performance on the development--calibration cell.

Calibration performance provides the error feedback used between rounds. Layer~3 is revised when calibration feedback favors an alternative mechanism, while the validated channel--evidence mappings remain fixed. Once a mechanism is retained, both are frozen for evaluation on unseen personas crossed with unseen products.

\section{Pricing study}
\label{sec:pricing}

The pricing study draws on Twin-2K-500, whose 40 purchase questions were asked in Wave~3 and re-asked in Wave~4. Each respondent--product case therefore has two human answers, and every prediction is scored separately against each. Because the two waves elicit the same items from the same respondents, their agreement quantifies intra-individual response stability and provides a reference ceiling for individual-level prediction. On the double holdout defined below, test--retest agreement is 85.0\% (95\% persona-cluster bootstrap interval [81.0, 88.5]), comparable to the 81.72\% mean across the 17 holdout tasks reported by \citet{toubia2025twin}. The study uses 50 randomly selected respondents and all 40 purchase questions. Before any label is read, a fixed-seed split (Figure~\ref{fig:data-split}) assigns 20 respondents to development and 30 to persona holdout. The 40 products are split within 10 consecutive groups of four---two to calibration, two to product holdout---yielding 20 calibration and 20 holdout products. Learning draws labels only from the development--calibration cell and the other three cells stay sealed, so the double holdout of $30\times20=600$ cases per wave contains only unseen respondents and unseen products. The primary pipeline simulator is \texttt{gpt-5.4-nano} (hereafter Nano) with high reasoning effort. For post-hoc capacity diagnostics, we run \texttt{gpt-5.4}, \texttt{gpt-5-mini}, and \texttt{qwen3-8b} on the same raw-transcript condition and double-holdout cell. The OpenAI comparators receive the archived Nano prompts verbatim, with only the model identifier changed. Each comparator is averaged over two simulator runs.\footnote{Resolved model snapshots: \texttt{gpt-5.4-nano-2026-03-17}, \texttt{gpt-5.4-2026-03-05}, and \texttt{gpt-5-mini-2025-08-07}.} Accuracy is the share of exact binary purchase predictions, averaged across repeated simulator runs.

\begin{figure}[!tbp]
  \centering
  \resizebox{\linewidth}{!}{%
  \begin{tikzpicture}[
    font=\sffamily\footnotesize,
    panel/.style={rounded corners=3pt, draw=dtmuted!55,
      line width=0.65pt, align=left, inner xsep=7pt, inner ysep=6pt,
      text width=6.35cm, minimum height=3.05cm}
  ]
    \node[panel, fill=dtbluefill] (history) at (3.30,0) {%
      \textbf{Non-target survey history (excerpt)}\\[3pt]
      \textbf{Q:} Do you have trouble limiting your spending?\\
      \textbf{Options:} Never / Rarely / Sometimes / Often / Always\\
      \textbf{A:} [respondent-specific response]\\[5pt]
      \textbf{Q:} The more commonplace a product or brand is among the
      general population, the less interested I am in buying it.\\
      \textbf{A:} [respondent-specific response]
    };
    \node[panel, fill=dtgreenfill] (target) at (10.55,0) {%
      \textbf{Target purchase item (representative)}\\[3pt]
      \textbf{Category:} Fruit Juice\\
      \textbf{Product:} Capri Sun Variety Pack with Fruit Punch,
      Strawberry Kiwi \& Pacific Cooler Juice Box Pouches, 30 ct Box,
      6 fl oz Pouches\\
      \textbf{Displayed price:} \$9.43\\[3pt]
      \textbf{Q:} Would you or would you not purchase this product?\\
      \textbf{Options:} Yes, I would purchase the product /
      No, I would not purchase the product\\
      \textbf{A:} [masked]
    };
  \end{tikzpicture}
  }
  \caption{\textbf{Pricing-task inputs.} Left: an excerpt of the non-target survey history from which persona evidence is selected. Right: a representative target purchase item with instrument wording. Respondent-specific answers and the target label are masked.
}
  \label{fig:pricing-input-example}
\end{figure}

\begin{figure}[!tbp]
  \centering
  \resizebox{0.84\linewidth}{!}{%
  \begin{tikzpicture}[
    font=\sffamily\scriptsize,
    cell/.style={rounded corners=2.5pt, minimum width=2.75cm,
      minimum height=0.76cm, align=center, line width=0.6pt},
    axis/.style={font=\sffamily\bfseries\scriptsize, text=dtink,
      align=center}
  ]
    \node[axis] at (4.00,1.64) {Calibration products (20)};
    \node[axis] at (7.05,1.64) {Holdout products (20)};
    \node[axis, anchor=east] at (2.24,0.78)
      {Development personas\\(20)};
    \node[axis, anchor=east] at (2.24,-0.18)
      {Holdout personas\\(30)};

    \node[cell, draw=dtorange, fill=dtorangefill, text=dtink]
      at (4.00,0.78) {\textbf{LEARNING CELL}\\labels guide updates};
    \node[cell, draw=dtgreen, fill=dtgreenfill, text=dtink]
      at (7.05,0.78) {\textbf{PRODUCT HOLDOUT}\\labels remain sealed};
    \node[cell, draw=dtgreen, fill=dtgreenfill, text=dtink]
      at (4.00,-0.18) {\textbf{PERSONA HOLDOUT}\\labels remain sealed};
    \node[cell, draw=dtgreen, fill=dtgreen!24, text=dtink,
      line width=0.9pt] at (7.05,-0.18)
      {\textbf{DOUBLE HOLDOUT}\\600 cases per wave};
  \end{tikzpicture}}
  \caption{\textbf{Two-axis data split.} Only the upper-left cell supplies learning labels. The double holdout contains unseen personas and products.}
  \label{fig:data-split}
\end{figure}

We compare two respondent representations, both constructed only from the non-target survey history. \raw{} supplies the original transcript of question--answer pairs (Figure~\ref{fig:pricing-input-example}). \cmech{} supplies the two learned channel definitions, persona-specific source evidence allocated to them, and the frozen executable mechanism, including transformations, scope restrictions, and the separate-routing policy. \cmech{} excludes the remaining raw transcript. Because \cmech{} derives its evidence from the source transcript, it changes evidence selection, compression, and routing rather than adding respondent facts. Appendix~\ref{app:frozen-pricing-artifact} gives a content-preserving rendering of the final auto-discovered prompt artifact.

\section{Results and interpretation}
\label{sec:results}

\subsection{Learned channels, mechanism, and discovery trace}
Figure~\ref{fig:calibration-trace} summarizes the training process as three rounds. In Round~1, Layer~1 proposed six candidate channels from the study description and the texts of the 20 calibration target questions, and Layer~2 allocated respondent-specific survey evidence to each candidate and evaluated the resulting mappings on the development--calibration cell. Round~1 separately evaluated and retained two channels---spending restraint and brand distinctiveness---whose persona representation draws on seven source survey questions in total: four for spending restraint and three for brand distinctiveness. Because these channels passed Layer~2 validation, the next round followed the affirmative branch in Figure~\ref{fig:pipeline} into Layer~3. Round~2 entered Layer~3 with the two channels retained in Round~1 and proposed an initial explicit-interaction mechanism for combining them. This condition outperformed the single-channel controls, but the comparison was not mechanism-specific: it changed both the number of channels used and the rule for combining them. Round~3 therefore held the two-channel representation fixed and varied only the mechanism. It compared the explicit interaction, which required a joint configuration of spending restraint and brand distinctiveness, with separate routing, which allowed each channel to influence the prediction independently under its own context gate. Separate routing achieved the higher development--calibration accuracy and was retained as the frozen mechanism.

\begin{figure}[!tbp]
  \centering
  \resizebox{\linewidth}{!}{%
  \begin{tikzpicture}[
    font=\sffamily,
    panel/.style={rounded corners=3pt, draw=dtmuted!55, fill=white,
      line width=0.65pt, minimum width=4.05cm, minimum height=5.35cm},
    candidate/.style={rounded corners=2.2pt, draw=dtblue!70,
      fill=dtbluefill, line width=0.65pt, text width=3.35cm,
      align=center, inner xsep=5pt, inner ysep=4pt, font=\scriptsize},
    chosen/.style={rounded corners=2.2pt, draw=dtgreen,
      fill=dtgreenfill, line width=0.85pt, text width=3.35cm,
      align=center, inner xsep=5pt, inner ysep=4pt, font=\scriptsize},
    rejected/.style={rounded corners=2.2pt, draw=dtorange,
      fill=dtorangefill, dashed, line width=0.75pt, text width=3.35cm,
      align=center, inner xsep=5pt, inner ysep=4pt, font=\scriptsize},
    flow/.style={-{Latex[length=2.1mm,width=1.55mm]}, draw=dtmuted,
      line width=0.8pt}
  ]
    \node[panel] (p1) at (2.10,2.68) {};
    \node[panel] (p2) at (6.55,2.68) {};
    \node[panel] (p3) at (11.00,2.68) {};

    \node[font=\bfseries\scriptsize, text=dtblue, align=center]
      at (2.10,4.83) {ROUND 1\\CHANNEL ABLATION};
    \node[candidate] at (2.10,3.55)
      {\textbf{Spending restraint only}\\[-1pt]
       Acc. $.7075$ \; ({\color{dtblue}\textbf{$+1.75$ pp}})};
    \node[candidate] at (2.10,2.25)
      {\textbf{Brand distinctiveness only}\\[-1pt]
       Acc. $.7075$ \; ({\color{dtblue}\textbf{$+1.75$ pp}})};
    \node[font=\scriptsize, text=dtmuted] at (2.10,1.18)
      {Same-round \raw{}: Acc. $.6900$};
    \node[font=\scriptsize\bfseries, text=dtgreen] at (2.10,0.52)
      {Retain 2 of 6 channels};

    \node[font=\bfseries\scriptsize, text=dtpurple, align=center]
      at (6.55,4.78) {ROUND 2\\TWO-CHANNEL\\REPRESENTATION};
    \node[candidate, draw=dtpurple!75, fill=dtpurplefill] at (6.55,3.38)
      {\textbf{Same-round controls}\\[2pt]
       Spending only: $.6913$ \; (\textbf{$+1.75$ pp})\\
       Brand only: $.6900$ \; (\textbf{$+1.63$ pp})};
    \node[candidate, draw=dtpurple!75, fill=dtpurplefill] at (6.55,1.82)
      {\textbf{Initial explicit interaction}\\[-1pt]
       {\scriptsize Joint trigger over the same evidence}\\[-1pt]
       Acc. $.7150$ \; (\textbf{$+4.13$ pp})};
    \node[font=\scriptsize, text=dtmuted] at (6.55,0.62)
      {Same-round \raw{}: Acc. $.6738$};

    \node[font=\bfseries\scriptsize, text=dtink, align=center]
      at (11.00,4.78) {ROUND 3\\MECHANISM\\ABLATION};
    \node[chosen] at (11.00,3.32)
      {\textbf{Separate routing} \; {\color{dtgreen}\textbf{SELECT}}\\[-1pt]
       {\scriptsize Separate context-gated shifts;\\no cross-channel term.}\\[-1pt]
       {\scriptsize Acc. $.7300$ \; (\textbf{$+6.50$ pp})}};
    \node[rejected] at (11.00,1.48)
      {\textbf{Explicit interaction} \; {\color{dtorange}\textbf{REJECT}}\\[-1pt]
       {\scriptsize High restraint $+$ commonplace brand\\requires stronger category relevance.}\\[-1pt]
       {\scriptsize Acc. $.7050$ \; (\textbf{$+4.00$ pp})}};
    \node[font=\scriptsize, text=dtmuted] at (11.00,0.28)
      {Same-round \raw{}: Acc. $.6650$};

    \draw[flow] (4.13,2.68) -- (4.52,2.68);
    \draw[flow] (8.58,2.68) -- (8.97,2.68);
  \end{tikzpicture}}
  \caption{\textbf{Auto-discovery rounds.} Round~1 retains two of six channels.
  Round~2 compares their initial interaction with same-round single-channel
  controls. Round~3 fixes the representation and selects separate routing over
  the interaction.}
  \label{fig:calibration-trace}
\end{figure}

\subsection{Double-holdout performance}
The frozen representation and mechanism improve the same simulator on unseen respondents and unseen products. Table~\ref{tab:pricing-results} reports double-holdout accuracy. Nano \raw{} conditions \texttt{gpt-5.4-nano} on the complete non-target survey transcript, whereas Nano \cmech{} replaces that transcript with the learned channel representation and frozen mechanism. Relative to Nano \raw{}, Nano \cmech{} improves accuracy by 5.00 percentage points in Wave~3 and 6.83 points in Wave~4. Both paired 95\% confidence intervals exclude zero. Relative to the human test--retest agreement of .850, Nano \cmech{} reaches 73\% and 74\% of this reference ceiling in Waves~3 and~4, compared with 67\% and 66\% for Nano \raw{}.

\begin{table}[!tbp]
  \centering
  \caption{\textbf{Double-holdout performance.} Exact accuracy on the
  30-persona, 20-product double holdout, with \texttt{gpt-5.4-nano} fixed as
  the simulator. Differences are paired against Nano \raw{} after within-persona
  averaging, with 95\% persona-cluster bootstrap intervals (200{,}000 resamples)
  in brackets. Human test--retest is the share of the same 600 cases in which the
  respondent gave the same answer in Wave~3 and Wave~4.}
  \label{tab:pricing-results}
  \vspace{3pt}
  {\scriptsize
  \setlength{\tabcolsep}{4.3pt}
  \begin{tabular*}{\linewidth}{@{\extracolsep{\fill}}lcccc@{}}
    \toprule
    & \multicolumn{2}{c}{Wave 3} & \multicolumn{2}{c}{Wave 4} \\
    \cmidrule(lr){2-3}\cmidrule(lr){4-5}
    \textbf{Simulator and input} & \textbf{Accuracy}
      & \shortstack{\textbf{$\Delta$ vs. Nano \raw{}}\\\textbf{[95\% CI]}}
      & \textbf{Accuracy}
      & \shortstack{\textbf{$\Delta$ vs. Nano \raw{}}\\\textbf{[95\% CI]}} \\
    \midrule
    Nano \raw{}                 & .5725 & --- & .5575 & --- \\
    Nano \cmech{}               & .6225 & +.0500 [.0150, .0858]
                                & .6258 & +.0683 [.0358, .1008] \\
    \midrule
    Human test--retest (W3 $\leftrightarrow$ W4) & \multicolumn{4}{c}{.8500 [.8100, .8850]} \\
    \bottomrule
  \end{tabular*}}
\end{table}

Table~\ref{tab:capacity-diagnostic} asks a narrower question: does any simulator reading the full \raw{} transcript exceed Nano equipped with the learned mechanism and routing? We reran three comparators on the double holdout cell under the archived \raw{} prompts and paired them, within persona, against the archived Nano \raw{} and Nano \cmech{} results. None of the six comparator--wave contrasts against Nano \cmech{} excludes zero: no \raw{} comparator detectably exceeds the learned representation in either wave. The gain is also consistent across waves: Nano \cmech{} significantly improves over Nano \raw{} in both waves (Table~\ref{tab:pricing-results}), a pattern no \raw{} comparator matches.

\begin{table}[!tbp]
  \centering
  \caption{\textbf{Post-hoc capacity diagnostic.} Exact accuracy on the
  30-persona, 20-product double holdout. All comparator simulators receive the
  \raw{} transcript. Entries report paired differences with 95\%
  persona-cluster bootstrap intervals (200{,}000 resamples) against the archived
  Nano \raw{} and Nano \cmech{} references, respectively.}
  \label{tab:capacity-diagnostic}
  \vspace{3pt}
  {\scriptsize
  \setlength{\tabcolsep}{3.5pt}
  \begin{tabular*}{\linewidth}{@{\extracolsep{\fill}}lccc@{}}
    \toprule
    \textbf{RAW simulator} & \textbf{Accuracy}
      & \shortstack{\textbf{$\Delta$ vs. Nano}\\\textbf{RAW}}
      & \shortstack{\textbf{$\Delta$ vs. Nano}\\\textbf{Channel+Mechanism}} \\
    \midrule
    \multicolumn{4}{@{}l}{\textbf{Wave 3}} \\
    \addlinespace[1pt]
    \texttt{gpt-5.4}      & .6050 & +.0325 [$-$.0042, .0700] & $-$.0175 [$-$.0658, .0308] \\
    \texttt{gpt-5-mini}   & .5858 & +.0133 [$-$.0342, .0617] & $-$.0367 [$-$.0892, .0150] \\
    \texttt{qwen3-8b}     & .6075 & +.0350 [$-$.0117, .0792] & $-$.0150 [$-$.0617, .0308] \\
    \addlinespace[2pt]
    \cmidrule(lr){1-4}
    \addlinespace[2pt]
    \multicolumn{4}{@{}l}{\textbf{Wave 4}} \\
    \addlinespace[1pt]
    \texttt{gpt-5.4}      & .6083 & +.0508 [.0150, .0867] & $-$.0175 [$-$.0700, .0342] \\
    \texttt{gpt-5-mini}   & .5892 & +.0317 [$-$.0108, .0733] & $-$.0367 [$-$.0908, .0150] \\
    \texttt{qwen3-8b}     & .5975 & +.0400 [$-$.0025, .0800] & $-$.0283 [$-$.0683, .0108] \\
    \bottomrule
  \end{tabular*}}
\end{table}

\subsection{What supports the gain}
Table~\ref{tab:layer-necessity} reports two post-hoc layer ablations and their combination. The original Layer~1 is answer-blind but target-aware: it sees the text of the 20 calibration target questions, including each product and price, but not the corresponding human answers. The Layer~1 ablation removes those target-question texts and the study description. From the answer-blind source-question catalog alone, it proposes six channel--anchor bundles. The Layer~2 ablation replaces the channels retained by Layer~2 with a pair that Layer~2 had measured as unsupported. In each ablation, all other pipeline components are held fixed.

Table~\ref{tab:layer-necessity} pairs each ablated pipeline directly against the full \cmech{} condition. Every ablation produces a statistically detectable loss: all six ablation--wave contrasts exclude zero. These post-hoc comparisons therefore support contributions from both target-aware channel discovery and Layer~2 validation and retention. Two additional contrasts clarify how these contributions depend on the state of the other layer. The Layer~1 row exceeds the joint row by $+.0469$ $[.0119,.0808]$ in Wave~3 and $+.0447$ $[.0117,.0778]$ in Wave~4, so restoring Layer~2 measurement and selection when Layer~1 remains target-blind yields a detectable recovery. By contrast, the Layer~2 row exceeds the joint row by $+.0297$ $[-.0061,.0661]$ and $+.0131$ $[-.0197,.0472]$, respectively, so the incremental contribution of target-aware Layer~1 when Layer~2 validation is already ablated is statistically unresolved.

\begin{table}[!tbp]
  \centering
  \caption{\textbf{Post-hoc layer ablations.} Exact accuracy on the
  30-persona, 20-product double holdout. Each row changes the named layer and holds the remainder fixed.
  Differences are paired against the full Nano \cmech{} condition, with 95\%
  persona-cluster bootstrap intervals in brackets.}
  \label{tab:layer-necessity}
  \vspace{3pt}
  {\scriptsize
  \setlength{\tabcolsep}{4pt}
  \begin{adjustbox}{max width=\linewidth}
  \begin{tabular}{@{}lcccc@{}}
    \toprule
    & \multicolumn{2}{c}{Wave 3} & \multicolumn{2}{c}{Wave 4} \\
    \cmidrule(lr){2-3}\cmidrule(lr){4-5}
    \textbf{Ablated layer} & \textbf{Accuracy}
      & \shortstack{\textbf{$\Delta$ vs. Nano}\\\textbf{\cmech{}}}
      & \textbf{Accuracy}
      & \shortstack{\textbf{$\Delta$ vs. Nano}\\\textbf{\cmech{}}} \\
    \midrule
    None: Nano \cmech{}           & .6225 & --- & .6258 & --- \\
    Layer~1, target text removed  & .5881 & $-$.0344 [$-$.0636, $-$.0053]
                                  & .5892 & $-$.0367 [$-$.0683, $-$.0033] \\
    Layer~2, evidence validation  & .5708 & $-$.0517 [$-$.0942, $-$.0067]
                                  & .5575 & $-$.0683 [$-$.1150, $-$.0175] \\
    Layers~1 and~2 together       & .5411 & $-$.0814 [$-$.1192, $-$.0428]
                                  & .5444 & $-$.0814 [$-$.1194, $-$.0417] \\
    \bottomrule
  \end{tabular}
  \end{adjustbox}}
\end{table}

\subsection{Decision-level interpretation}
The learned representation makes three parts of a simulated decision inspectable: the respondent-specific survey evidence assigned to each channel, the product context that makes a channel relevant, and the direction in which that evidence may shift the prediction. Figure~\ref{fig:case-trace} traces one such case. Each channel summary is synthesized from the same anonymized respondent’s non-target answers while preserving conflicting evidence. The resulting trace provides an auditable evidence--context--prediction mapping at the input--output level, but does not establish that the textual mechanism reflects the model’s internal computation or the respondent’s causal psychological process.

The learned channels and routing are also consistent with established consumer-behavior accounts. Spending restraint routed by price and necessity is consistent with mental budgeting and self-control accounts that emphasize context-sensitive spending constraints~\citep{heath1996mental,hoch1991selfcontrol}. Brand distinctiveness gated by a mainstream--distinctive interpretation is consistent with need for uniqueness, which concerns attraction to differentiated products and brands rather than general spending propensity~\citep{tian2001uniqueness}. Its application across product categories is also compatible with evidence of correlated consumer sensitivities across categories~\citep{ainslie1998crosscategory}. These connections support an interpretive reading of the learned channels, but do not show that the LLM recovered the respondent's psychological process or that the textual rule is causal.

\begin{figure}[!tbp]
  \centering
  \resizebox{\linewidth}{!}{%
  \begin{tikzpicture}[
    font=\sffamily\scriptsize,
    rawanswer/.style={rounded corners=3pt, draw=dtblue!72, fill=dtbluefill!58,
      line width=0.65pt, align=left, inner xsep=6pt, inner ysep=5pt,
      text width=2.90cm, minimum height=1.45cm, text=dtink},
    evidence/.style={rounded corners=3pt, draw=dtblue, fill=dtbluefill,
      line width=0.7pt, align=left, inner xsep=6pt, inner ysep=5pt,
      text width=2.25cm, minimum height=1.45cm, text=dtink},
    product/.style={rounded corners=3pt, draw=dtpurple, fill=dtpurplefill,
      line width=0.7pt, align=left, inner xsep=7pt, inner ysep=5pt,
      text width=4.10cm, minimum height=1.08cm, text=dtink},
    router/.style={rounded corners=3pt, draw=dtgreen, fill=dtgreenfill,
      line width=0.8pt, align=left, inner xsep=8pt, inner ysep=6pt,
      text width=3.95cm, minimum height=2.72cm, text=dtink},
    decision/.style={rounded corners=3pt, draw=dtmuted!65, fill=white,
      line width=0.8pt, align=center, inner xsep=8pt, inner ysep=6pt,
      text width=2.40cm, minimum height=2.85cm, text=dtink},
    flow/.style={-{Latex[length=2.1mm,width=1.55mm]}, draw=dtmuted,
      line width=0.8pt, rounded corners=3pt}
  ]
    \node[font=\sffamily\scriptsize\bfseries, text=dtblue, align=center]
      at (1.65,2.65) {1. RAW SURVEY ANSWERS};
    \node[rawanswer] (rawbrand) at (1.65,1.15) {%
      \textbf{Brand source answers}\\[2pt]
      Avoid popular brands: \textbf{Strongly agree}\\
      Commonplace brands reduce interest: \textbf{Strongly agree}
    };
    \node[rawanswer] (rawspending) at (1.65,-1.25) {%
      \textbf{Spending source answers}\\[2pt]
      Tightwad--spendthrift: \textbf{8/11 toward spendthrift}\\
      Trouble limiting spending: \textbf{Rarely}\\
      Shopper scenario: \textbf{4/5 toward tightwad}
    };

    \node[font=\sffamily\scriptsize\bfseries, text=dtblue, align=center]
      at (4.80,2.65) {2. LAYER-2 CHANNEL\\SYNTHESIS};
    \node[evidence] (brand) at (4.80,1.15) {%
      \textbf{Brand distinctiveness}\\[2pt]
      Strong mainstream-brand avoidance
    };
    \node[evidence] (spending) at (4.80,-1.25) {%
      \textbf{Spending restraint}\\[2pt]
      Mixed spending signals; no clear restraint level
    };

    \node[product] (context) at (8.65,2.75) {%
      \textbf{3. UNSEEN PRODUCT}\\[2pt]
      Great Value chicken wings, 4 lb; \textbf{\$12.98}\\
      Brand cue shown; category need not supplied
    };

    \node[router] (route) at (8.65,-0.55) {%
      \textbf{4. LAYER-3 FROZEN}\\[-1pt]
      \textbf{SEPARATE ROUTING}\\[4pt]
      \textbf{Brand route:} if ``Great Value'' is classified as mainstream,
      strong mainstream-brand avoidance shifts toward \textbf{No}.\\[4pt]
      \textbf{Spending route:} Mixed spending evidence supports no directional shift; missing evidence about category need is treated as neutral. \\[3pt]
      \textbf{Separate routing:} retain only eligible directional shifts.
    };

    \node[decision] (answer) at (12.95,-0.55) {%
      \textbf{5. FINAL ANSWER}\\[6pt]
      Nano\\[-1pt]
      Channel + Mechanism: \\[-1pt]
      \textcolor{dtgreen}{\Large\textbf{No}}\\[5pt]
      Nano \raw{}: \textbf{Yes}\\
      Human Wave~3: \textbf{No}
    };

    \draw[flow, draw=dtblue] (rawbrand.east) -- (brand.west);
    \draw[flow, draw=dtblue] (rawspending.east) -- (spending.west);
    \draw[flow] (brand.east) to[out=0,in=180] ([yshift=0.62cm]route.west);
    \draw[flow] (spending.east) to[out=0,in=180] ([yshift=-0.62cm]route.west);
    \draw[flow, draw=dtpurple] (context.south) -- (route.north);
    \draw[flow, draw=dtgreen] (route.east) -- (answer.west);
  \end{tikzpicture}}
  \caption{\textbf{Decision-level trace under the separate-routing mechanism.} Post-hoc example for one anonymized persona and one holdout product.
  Representative source items are paraphrased. Each Layer-2 channel synthesizes
  that respondent's own non-target answers while preserving conflicts.}
  \label{fig:case-trace}
\end{figure}

\subsection{Scope and limitations}
The evaluation establishes system-level transfer across unseen respondents and unseen products within a common binary purchase template. It supports transfer within the pricing task, not across survey constructs or task formats. Appendix~\ref{app:signal-diagnostics} adds a boundary diagnostic: the same evidence and channel construction applied to a cognitive-bias battery exposes no detectable persona-specific signal under the present design, which bounds the result rather than extending it.

The main methodological limitation concerns downstream rule capacity. The selected Layer-3 mechanism is a simple context-gated separate-routing textual rule. The supervised diagnostic in Appendix~\ref{app:signal-diagnostics} performs better using the same upstream channel representation, suggesting that the current textual mechanism does not fully exploit the available predictive signal. Because the diagnostic receives direct human calibration labels, it should be interpreted as a capacity probe rather than a like-for-like replacement. Nonetheless, even the learned representation remains well below the .850 human test--retest agreement on the double holdout. The reported gains therefore constitute relative improvements over the transcript baseline, not evidence that simulated responses approach the stability of the respondents' own repeated answers.

Future work should compare more expressive rule learners and test whether language models with different capacities or inductive biases discover stronger mechanisms from the same calibration feedback. Any added complexity should be selected before the double-holdout evaluation, verified on unseen personas and products, and subsequently tested in independent studies with different decision structures.

\section{Conclusion}

The auto-discovery pipeline separates what behavioral channels to represent, which transcript answers instantiate them, and how that evidence should affect a prediction. In the pricing study, this evidence-to-decision mapping is made explicit through a context-gated decision rule rather than left implicit in the simulator. The resulting two-channel representation and frozen mechanism improve Nano accuracy over the full transcript on unseen respondents and products in both survey waves, while none of the tested RAW-transcript comparators detectably exceeds this condition in either wave. Post-hoc ablations further show detectable losses when target-aware channel discovery or evidence validation is removed. Together, these results show that task-specific evidence selection and routing can support within-task transfer while preserving an auditable evidence–context–prediction mapping. The evidence is limited to the pricing task and does not establish cross-task generalization, model equivalence or superiority, an interpretation of internal computation, or a causal account of human choice.

{\small
\setlength{\bibsep}{1pt}
\bibliographystyle{plainnat}
\bibliography{references}
}

\appendix
\section{Frozen auto-discovered pricing artifact}
\label{app:frozen-pricing-artifact}

The final artifact has two parts. One part holds the shared channel semantics
and routing instructions. The other holds respondent-specific evidence. Every
respondent receives the same shared text. Only the \texttt{source\_evidence}
fields change. Layer~2 fills those fields with the question and answer from each
available non-target source item.
Figure~\ref{fig:frozen-pricing-artifact} keeps the content of the frozen holdout
prompt but replaces every respondent answer with an angle-bracket placeholder.
The figure is an input-schema example rather than a model-generated explanation
of any particular prediction.

\begin{figure}[!tbp]
  \centering
  \fbox{\begin{minipage}{0.94\linewidth}
  \small
  \textbf{Pricing evidence router (answer-blind).}
  Treat the extracted evidence as soft, falsifiable priors rather than known
  motives or target answers. For each product, route only relevant channels;
  preserve counterevidence and conflicts. Missing category need is neutral,
  and silence does not establish pet ownership, medical need, diet, alcohol
  use, or a fixed purchase response.

  \vspace{4pt}
  \textbf{Channel 1: \texttt{spending\_restraint}}\\
  \emph{Definition:} general ease or discomfort with discretionary spending.\\
  \emph{Respondent-specific source evidence:}
  \begin{itemize}
    \setlength{\itemsep}{0pt}\setlength{\parskip}{0pt}
    \item Spending habits $\rightarrow$ \texttt{<this respondent's answer>}
    \item Trouble limiting spending $\rightarrow$ \texttt{<answer>}
    \item Trouble spending money $\rightarrow$ \texttt{<answer>}
    \item Tightwad--spendthrift self-comparison $\rightarrow$ \texttt{<answer>}
  \end{itemize}
  \emph{Transformation and scope:} stronger restraint raises the purchase
  threshold, especially for discretionary products; spendthrift tendencies
  may lower it. Preserve conflicts. Apply this as global price sensitivity
  moderated by product necessity, not as a fixed ``No.''

  \vspace{4pt}
  \textbf{Channel 2: \texttt{brand\_distinctiveness}}\\
  \emph{Definition:} preference for distinctive versus mainstream products
  and brands.\\
  \emph{Respondent-specific source evidence:}
  \begin{itemize}
    \setlength{\itemsep}{0pt}\setlength{\parskip}{0pt}
    \item Buying special products or brands to develop uniqueness
      $\rightarrow$ \texttt{<answer>}
    \item Avoiding products or brands bought by the general population
      $\rightarrow$ \texttt{<answer>}
    \item Reduced interest as a product or brand becomes commonplace
      $\rightarrow$ \texttt{<answer>}
  \end{itemize}
  \emph{Transformation and scope:} strong distinctiveness seeking may reduce
  attraction to commonplace brands and increase attraction to unusual
  products. It does not imply willingness to pay. Apply it only to
  brand/product identity, moderated by category need and price.

  \vspace{4pt}
  \textbf{Frozen additive mechanism.}
  Apply spending restraint and brand distinctiveness separately. Spending
  restraint may change the general purchase threshold; brand distinctiveness
  may affect attraction to a clearly mainstream brand. Do not create a joint
  trigger, joint penalty, multiplicative effect, or requirement that both
  conditions be present.
  \end{minipage}}
  \caption{\textbf{Frozen simulator artifact.} Content-preserving example of the
  \cmech{} artifact supplied to the simulator. All respondents share the same
  channel definitions, transformations, scope restrictions, and archived
  ``additive'' mechanism. The main text calls that mechanism separate routing.
  It implies no numerical addition and no causal additivity. Each respondent's
  own non-target survey answers fill the seven \texttt{source\_evidence} values.}
  \label{fig:frozen-pricing-artifact}
\end{figure}

\section{Diagnosing the three-layer pipeline}
\label{app:signal-diagnostics}
The main text defines three learned objects. Layer~1 reads the target
calibration questions and derives a channel skeleton from the constructs,
contexts, and decision demands they express. This step fixes what the
representation may need to contain. Layer~2 instantiates those channels for each
respondent. It allocates non-target survey answers to every channel, keeping
counterevidence and uncertainty intact. It then validates each
channel--evidence mapping through a controlled exposure intervention. Layer~3
turns the validated channel values into an executable consumer mechanism. That
mechanism states when the evidence applies and in what direction it should move
a prediction. It also states which moderators, interactions, and prohibited
scopes constrain that use.

The following post-hoc analyses diagnose the Section~\ref{sec:results} result.
They are not additional contributions. They serve two purposes. The
cognitive-bias analyses are boundary diagnostics. They ask whether the same
evidence and channel construction exposes detectable persona-specific signal
outside pricing. The pricing decision-rule diagnostic probes downstream rule
capacity. It holds the Layer-1 and Layer-2 representation fixed and asks whether
a different low-complexity rule can exploit further persona-specific signal.
None of these experiments identifies the three layers' contributions separately
unless the stated comparison holds the other learned objects fixed. The
accompanying tables report every empirical comparison.

\paragraph{Cognitive-bias boundary diagnostic: channel-versus-Empty intervention.}
We compare each frozen task-routed channel with a shared Empty condition over
the complete cognitive-bias battery. The primary \texttt{gpt-5.4-nano} simulator
runs every condition. Empty contains the questions and the response schema but
no persona-specific information. Each channel condition contains only its own
answer-blind source evidence and frozen router. It never includes the full raw
transcript. Table~\ref{tab:cognitive-channel-full-battery} reports two repeated
generations and persona-level paired contrasts. These cover the calibration
responses and the subsequent Wave~4 responses.

\begin{table}[!tbp]
  \centering
  \caption{\textbf{Cognitive-bias channel diagnostic.} Complete-battery task-balanced normalized accuracy (\%) and
  channel-minus-Empty contrasts for the cognitive-bias diagnostic. Empty and
  channel conditions each have two runs. Contrasts compare their means with a
  persona-level paired bootstrap over 20{,}000 resamples. No adjustment is made
  across the nine channels. Every row has $n{=}20$ apart from public reference
  frame at $n{=}18$. That row therefore has its own Empty baseline.}
  \label{tab:cognitive-channel-full-battery}
  \vspace{3pt}
  {\scriptsize
  \setlength{\tabcolsep}{3pt}
  \begin{adjustbox}{max width=\linewidth}
  \begin{tabular}{@{}lrrrrrrr@{}}
    \toprule
    \textbf{Channel} & \textbf{$n$} & \textbf{Empty 1} & \textbf{Empty 2}
      & \textbf{Ch. 1} & \textbf{Ch. 2}
      & \shortstack{\textbf{Cal. $\Delta$}\\\textbf{[95\% CI]}}
      & \shortstack{\textbf{Wave 4 $\Delta$}\\\textbf{[95\% CI]}} \\
    \midrule
    Conditional directionality & 20 & 68.38 & 69.64 & 68.44 & 69.11
      & $-0.23\ [-2.42,+2.08]$ & $-1.36\ [-3.15,+0.34]$ \\
    Environmental projection & 20 & 68.38 & 69.64 & 69.61 & 70.98
      & $+1.29\ [-1.10,+3.60]$ & $+1.39\ [-0.93,+3.82]$ \\
    Probability translation & 20 & 68.38 & 69.64 & 69.52 & 69.05
      & $+0.27\ [-2.10,+2.39]$ & $-0.38\ [-2.90,+2.04]$ \\
    Gain-lottery CE & 20 & 68.38 & 69.64 & 69.44 & 69.43
      & $+0.42\ [-1.51,+2.35]$ & $+0.09\ [-2.68,+2.67]$ \\
    Mental-accounting integration & 20 & 68.38 & 69.64 & 67.42 & 69.56
      & $-0.52\ [-2.78,+1.49]$ & $-0.84\ [-3.16,+1.61]$ \\
    Numeric magnitude estimation & 20 & 68.38 & 69.64 & 68.83 & 68.31
      & $-0.44\ [-2.35,+1.39]$ & $-1.10\ [-3.91,+1.39]$ \\
    Public reference frame & 18 & 67.81 & 69.47 & 68.73 & 68.42
      & $-0.07\ [-2.57,+2.22]$ & $+1.00\ [-1.02,+3.00]$ \\
    Social-vignette diagnosticity & 20 & 68.38 & 69.64 & 67.84 & 68.44
      & $-0.87\ [-3.10,+1.38]$ & $-1.42\ [-4.25,+1.29]$ \\
    Small-risk resolution & 20 & 68.38 & 69.64 & 69.80 & 70.95
      & $+1.36\ [-1.10,+3.72]$ & $-0.26\ [-2.87,+2.29]$ \\
    \bottomrule
  \end{tabular}
  \end{adjustbox}}
\end{table}

This intervention adds three things at once relative to Empty. It supplies the
proposed channel, its allocated evidence, and its task router. The test
therefore covers the combined output of Layers~1--3. It cannot separate channel
content from evidence allocation, routing, salience, or the executable
decision mechanism.

\paragraph{Pricing decision-rule diagnostic: downstream rule capacity.}
This diagnostic probes Layer~3 while the upstream representation stays fixed. We
freeze the pricing pipeline's two Layer-1 channels of \emph{spending restraint}
and \emph{brand distinctiveness}. We also freeze the Layer-2 source answers
allocated to them. Shallow regression trees then fit human calibration outcomes.
One tree reads the source answers directly. Another reads the normalized channel
scores. Both receive the same product-side features. Each tree is an alternative
low-complexity mapping from fixed persona evidence and product context to a
response. Table~\ref{tab:tree-diagnostic} reports the selected specifications and
the double-holdout results next to the frozen LLM reference.

\begin{table}[!tbp]
  \centering
  \caption{\textbf{Pricing decision-rule diagnostic.} Results with fixed Layer-1
  channels and Layer-2 evidence. All rows are scored against Wave~4 human
  responses. Tree specifications are selected on the 20 development respondents
  and 20 calibration products. They are then evaluated on the 30 unseen
  respondents and 20 unseen products. Every tree also receives 14 product
  features. Shuffled accuracy averages 200 persona reassignments. Paired
  intervals use 20{,}000 resamples. The seven-answer tree and its shuffled
  control emit identical binary predictions on the holdout. Their paired accuracy
  difference is therefore identically zero even though their squared and absolute
  errors differ.}
  \label{tab:tree-diagnostic}
  \vspace{3pt}
  {\scriptsize
  \setlength{\tabcolsep}{4pt}
  \begin{adjustbox}{max width=\linewidth}
  \begin{tabular}{@{}llcccc@{}}
    \toprule
    \textbf{Predictor} & \textbf{Persona features} & \textbf{Depth/leaf}
      & \textbf{Accuracy} & \shortstack{\textbf{Shuffled}\\\textbf{accuracy}}
      & \shortstack{\textbf{Assigned--shuffled $\Delta$}\\\textbf{[95\% CI]}} \\
    \midrule
    LLM simulator & Two channel prompts & --- & .6258 & --- & --- \\
    Decision tree & Seven source answers & $3/40$ & .6583 & .6583
      & $.0000\ [.0000,.0000]$ \\
    Decision tree & Two channel scores & $3/10$ & .6850 & .6517
      & $+.0333\ [+.0100,+.0600]$ \\
    \bottomrule
  \end{tabular}
  \end{adjustbox}}
\end{table}

The trees train directly on human calibration labels. This is therefore not a
like-for-like model comparison. What it probes is whether the frozen Layer-1
channels and Layer-2 evidence expose low-dimensional persona-specific signal to
a supervised decision rule. Aggregation is itself part of the learned
representation. We therefore evaluate the raw source answers separately from the
two aggregated channel scores. The result sets no general reasoning limit for
LLMs. It benchmarks the signal reachable under the selected representation,
model, and development sample.

\paragraph{Cognitive-bias boundary diagnostic: evidence-to-rule trees.}
A last test asks whether the cognitive-bias channels expose a low-dimensional
persona signal when no transfer between questions is required. Each calibration
question gets its own shallow regression tree. That tree fits the normalized
human responses of the 20 development respondents. It is then evaluated on the
same question for the 30 persona-holdout respondents. The tree receives either
the seven channel means available as numeric features, the 28 answer-blind
source responses, or both. This same-question protocol isolates cross-persona
prediction. The tree never receives question features and is never evaluated on
a question other than the one it trained on. Development-only cross-validation
selects maximum depth~1 for every feature set and split at minimum leaf size 5
or 7. Table~\ref{tab:cognitive-tree-diagnostic} reports the resulting accuracies
and the contrasts against the development-persona question mean.

\begin{table}[!tbp]
  \centering
  \caption{\textbf{Cognitive-bias decision-rule diagnostic.} Same-question
  persona-holdout results. Accuracy is one minus absolute error on a $[0,1]$
  response scale. MSE uses the same scale. Results are task-balanced over 14
  families and five label-blind calibration-question splits. Contrasts compare
  each tree with the development-persona question mean. Paired respondent
  bootstrap intervals use 20{,}000 resamples at $n{=}30$.}
  \label{tab:cognitive-tree-diagnostic}
  \vspace{3pt}
  {\scriptsize
  \setlength{\tabcolsep}{3pt}
  \begin{tabular*}{\linewidth}{@{\extracolsep{\fill}}lccccc@{}}
    \toprule
    \textbf{Persona features} & \textbf{Tree acc.}
      & \shortstack{\textbf{Question-mean}\\\textbf{accuracy}}
      & \shortstack{\textbf{$\Delta$ accuracy (pp)}\\\textbf{[95\% CI]}}
      & \textbf{Tree MSE}
      & \shortstack{\textbf{Question-mean}\\\textbf{MSE}} \\
    \midrule
    Seven channel means & .7462 & .7489
      & $-0.27\ [-0.62,+0.06]$ & .1041 & .0996 \\
    28 source answers & .7445 & .7489
      & $-0.44\ [-0.62,-0.24]$ & .1034 & .0996 \\
    Answers $+$ means & .7452 & .7489
      & $-0.37\ [-0.77,+0.04]$ & .1054 & .0996 \\
    \bottomrule
  \end{tabular*}}
\end{table}

A task-level explanation remains open. Pricing may simply admit stable
low-dimensional decision rules. The simulator's gain would then reflect access
to a readily compressible signal rather than a distinctive LLM reasoning
capability. Cognitive bias behaves differently. Neither the channel intervention
nor the same-question trees recover a detectable improvement from the current
evidence and development split. That outcome does not show the survey to be free
of predictive information. Richer nonlinear structure remains untested. So do
alternative Layer-1 channels, Layer-2 evidence, and larger development
samples. What the outcome does show is narrower. The pipeline's failure here
cannot be blamed solely on an LLM that failed to exploit an otherwise detectable
shallow-tree rule.

\end{document}